**Open-Source Manually Annotated Vocal Tract Database for Automatic Segmentation from 3D MRI Using Deep Learning: Benchmarking 2D and 3D Convolutional and Transformer Networks**

*Subin Erattakulangara[1], Karthika Kelat[1], Katie Burnham[2], Rachel Balbi[2], Sarah E. Gerard[1], David Meyer[3,4,2], Sajan Goud Lingala[1,5]*

[1]*Roy J. Carver Department of Biomedical Engineering, University of Iowa, Iowa City, USA*
[2]*Janette Ogg Voice Research Center, Shenandoah University, Winchester, VA, USA*
[3]*School of Music, University of Iowa, Iowa City, USA*
[4]*Department of Communication Sciences and Disorders, University of Iowa, Iowa City, USA*
[5]*Department of Radiology, University of Iowa, Iowa City, USA*

Corresponding author
Sajan Goud Lingala, Ph.D
Assistant Professor
Roy J. Carver Department of Biomedical Engineering
Department of Radiology
University of Iowa
sajangoud-lingala@uiowa.edu

**ABSTRACT:**

*Objectives:* Accurate segmentation of the vocal tract from MRI data is essential for various voice, speech, and singing applications. Manual segmentation is time-intensive and susceptible to errors. This study aimed to evaluate the efficacy of deep learning algorithms for automatic vocal tract segmentation from 3D MRI.

*Study Design:* This study employed a comparative design, evaluating four deep learning architectures for vocal tract segmentation using an open-source dataset of 3D MRI scans of French speakers.

Methods: Fifty-three vocal tract volumes from 10 French speakers were manually annotated by an expert vocologist, assisted by two graduate students in voice science. These included 21 unique French phonemes and 3 unique voiceless tasks. Four state-of-the-art deep learning segmentation algorithms were evaluated: 2D slice-by-slice U-Net, 3D U-Net, 3D U-Net with transfer learning (pre-trained on lung CT), and 3D transformer U-Net (3D U-NetR). The STAPLE algorithm, which combines segmentations from multiple annotators to generate a probabilistic estimate of the true segmentation, was used to create reference segmentations for evaluation. Model performance was assessed using the Dice coefficient, Hausdorff distance, and structural similarity index measure.

Results: The 3D U-Net and 3D U-Net with transfer learning models achieved the highest Dice coefficients (0.896 ± 0.05 and 0.896 ± 0.04, respectively). The 3D U-Net with transfer learning performed comparably to the 3D U-Net while using less than half the training data. It, along with the 2D slice-by-slice U-Net models demonstrated lower variability in HD distance compared to the 3D U-Net and 3D U-NetR models. All models exhibited challenges in segmenting certain sounds, particularly /kõn/. Qualitative assessment by a voice expert revealed anatomically correct segmentations in the oropharyngeal and laryngopharyngeal spaces for all models except the 2D slice-by-slice UNET, and frequent errors with all models near bony regions (eg. teeth).

*Conclusions:* This study emphasizes the effectiveness of 3D convolutional networks, especially with transfer learning, for automatic vocal tract segmentation from 3D MRI. Future research should focus on improving the segmentation of challenging vocal tract configurations and refining boundary delineations.

**INTRODUCTION:**

Vocal production involves intricate modulations of the human vocal tract's shape by various articulators, including the lips, velum, and tongue. To analyze these shape changes, (e.g. in normal and impaired speech, or in singing) researchers have increasingly utilized modalities including electro-magnetic articulography (EMA)[1], endoscopy[2], projection X-ray[3], X-ray computed tomography[4], ultrasound[5], and magnetic resonance imaging (MRI)[6,7]. Amongst these, MRI has emerged as a powerful tool due to its non-invasive nature, flexibility in imaging orientation, ability to image deep vocal structures, and suitability for longitudinal studies[8,9]. However, MRI is relatively slow compared to other modalities, potentially prolonging data acquisition. Recent advancements in accelerated MRI techniques, such as parallel imaging with custom neck coils and compressed sensing, have enabled rapid imaging of the entire vocal tract during sustained speech sounds, completing volumetric scans in under 15 seconds within one exhalation[10–13]. These accelerated protocols have facilitated the creation of open-source volumetric scans of the vocal tract during sustained speech. Notable examples include a 10-speaker database producing various French language phonemes[13] and a 75-speaker database producing several English language vowels and consonants[11].

Quantitative assessment of vocal tract posture modulation, such as through 3D vocal tract area functions and 2D mid-sagittal vocal-tract area change analysis, has informed numerous voice and speech applications. These include understanding speaker-to-speaker variability[14–16], developing and optimizing vocal tract models for voice synthesis[17,18], and conducting acoustic and aerodynamic studies of 3D-printed vocal tracts derived from MRI data[19–21]. Accurate segmentation of the vocal tract from MRI data is crucial for these analyses. The vocal tract airspace in 3D MR images appears opaque in contrast to the gray soft tissues. To access vocal tract posture, the soft-tissues and the comprised airspace are typically segmented via manual annotation. However, this method is time and labor-intensive as well as error-prone. Producing an accurate vocal tract segmentation may require 90 minutes (or more) of manual editing, and studies of manually segmented vocal tracts typically have small sample sizes and therefore weak statistical power.

Early segmentation approaches relied on methods like region growing[22–24], active appearance models (AAMs)[25,26], level set[27], model-based template methods[28,29]. These methods often require significant manual intervention or are limited in their ability to handle the complex anatomical variations across individuals and different imaging protocols. For example, region growing methods rely on manual initialization of seed points, and often warrants manual correction of segmentations at spatial locations with low signal to noise, low resolution, and fuzzy boundaries separating soft-tissues and air-space. AAMs, level set, and template methods rely on manual identification of segmentation boundaries in a template image, and allow them to be deformed across the remaining images. It has largely been applied on 2D plus time data, and not on 3D volumetric vocal tract data due to difficulty in characterizing complex 3D vocal tract anatomy by using a single 2D template slice.

Recent advancements in deep learning have led to the development of automatic segmentation methods on 2D mid-sagittal dynamic MRI data during free-running speech, and on 3D volumetric vocal tract MRI data during sustained speech[30–39]. These methods were applied to either segment vocal-tract airspace, pharyngeal airspace, or segment soft-tissue structures (eg. lips, tongue, velum). Ruthven et al. developed custom 2D U-Net architectures for segmentation of vocal tract structures (head, velum, jaw, tongue, tooth space) on 2D plus time mid-sagittal data, and have also released ground-truth segmentations from five healthy volunteers as open source[39]. Erattakulangara et al. developed 2D U-Net architecture for vocal tract segmentation in mid-sagittal plane of a 3D volume, and utilized training with 75 slices across 10 speakers[31]. Liping Xie et al. developed 2D U-Net architectures with custom loss functions for a multitude of upper-airway MR segmentation tasks involving different kinds of datasets[35,36]. These included segmentation of sections of upper-airway relevant in sleep apnea (eg. oro-pharyngeal, hypopharyngeal, supraglottic/glottic airspace) from static 3D MRI; segmentation of deforming pharyngeal air-space in both 2D plus time mid-sagittal, and 3D plus time volumetric dynamic data. They also leveraged local neighborhood contextual relations by processing a slice of interest along with its two immediate spatial/or temporal slices (on either side) to predict its segmentation, and used a large, diverse database of upper-airway MR images to train their models (eg. up to 30 subjects with upto 7544 slices). Bommineni et al. used 2D U-Net architectures to segment 10 obstructive sleep apnea (OSA) relevant anatomical structures in 3D T1-weighted MRI in static posture, and automatically quantified risk factors in OSA (eg. size of soft-palate, volume of tongue-fat). Their study incorporated datasets from 234 participants enrolled in various clinical sleep studies, and after data-curation, their deep learning model employed data from 206 participants for training and testing[38]. To address efficient training with limited annotated samples, Erattakulangara et al. developed small data-based transfer learning 2D-U-Net models to segment tongue, velum, and vocal-tract airspace from mid-sagittal 2D plus time dynamic datasets[32]. This study showed reliable segmentations across three different MRI acquisition and reconstruction protocols with training sizes as few as 20 annotated samples.

Despite the progress made in deep learning methods for upper airway MRI segmentation, several factors limit the widespread applicability of these methods in 3D vocal tract segmentation from volumetric dynamic MRI data during sustained speech. First, current deep-learning models for 3D upper-airway MR segmentation require large amounts of labeled training data (eg. few 1000's of slices) [35,38]. Moreover, the data used in these models was taken from sleep studies where the target region of interest is the pharyngeal air-space relevant in OSA. This region of interest does not encompass the entire vocal tract airspace (beginning at the glottis and terminating at the lips). Second, due to the niche nature of upper airway research, there is a scarcity of open-source labelled upper-airway segmentation datasets, making it challenging and time-consuming to adapt current models to other applications. To the best of our knowledge, only the work of Ruthven et al. has open-source labeled vocal tract and articulators on mid-sagittal 2D plus time MRI data on 5 English speakers[39], and the work of Birkholz et al. has 3D vocal tract shapes from 2 speakers sustaining German speech sounds[19]. There are no other open-source annotated vocal tract volumes from 3D-MRI data during sustained voicing. Third, the majority of current models employ 2D convolutional neural networks to segment 3D volumes by slicing them into multiple frames, or by processing them along with immediate neighbors. The use of 2D convolutions cannot fully leverage features or relationships in 3D space, potentially leading to non-anatomical and inconsistent segmentations. To address these challenges, we make the following contributions in this paper:

1. We develop an open-source manually labelled 3D vocal tract database from the French speaker database[13]. We provide annotations of 53 vocal tract volumes (or 1696 2D slices with 32 slices per volume) manually segmented from 10 speakers producing different French phonemes and voiceless tasks. A total of 21 unique phonemes and 3 unique voiceless tasks were included. These segmentations were performed by a team of vocologists with expertise in vocal anatomy, supervised by one of the authors (D. Meyer). The resulting database adds diversity in the field of speech and voice science which is largely dominated by English speakers.
2. We evaluate the performance of four state-of-the-art deep learning segmentation algorithms, including:
   a) the slice-by-slice 2D U-Net method employing 2D convolutions;
   b) the 3D U-Net method employing 3D convolutions;
   c) the 3D U-Net method leveraging transfer learning by pre-training with open-source annotated samples from another lung CT imaging application;
   d) the 3D transformer U-Net model.

These models were chosen because they represent the state of the art in upper-airway MR segmentation literature (including the slice-by-slice 2D U-Net method and transfer learning models) and also serve to benchmark the utility of 3D models. Specifically, the 3D U-Net, which relies on 3D convolutions to leverage 3D spatial features, has been shown to be superior to the 2D slice-by-slice U-Net in other medical image segmentation tasks. Additionally, the 3D UNetR model employs vision transformers for pattern recognition instead of conventional convolutions. This approach has the potential to capture long-distance relationships between features in a 3D volume, though it has not yet been demonstrated in upper-airway MRI segmentation.

3. We quantitatively evaluate segmentations from the above algorithms against ground truth segmentations created by the simultaneous truth and performance level estimation (STAPLE) method[40], which mitigates inter-user human variability while creating ground truth segmentations. In this work, we used manual segmentations by 3 human experts who created them independently using the Slicer software and, computed a STAPLE probabilistic estimate of the true segmentation.
4. We provide open-source code of the above four models to facilitate ease of reproducibility by other researchers.

**METHODS:**

***Datasets and Pre-processing:*** Datasets used in this work were divided into two parts: the pre-training dataset, and the training dataset. The pre-training dataset contains data that were used for pre-training the 3D U-NET transfer learning network, and is sourced from the publicly available Pulmonary Fibrosis Competition dataset at the Open-Source Imaging Consortium (OSIC)[41]. This dataset includes chest CT scans and associated clinical information for a set of patients. A subset of this dataset (110 volumes) was manually segmented by Konya et al[42]. These segmentations included segmentations of the lungs, heart, and the trachea. 40 volumes and their corresponding lung segmentations from this dataset were randomly selected for this study. For training, we utilized a multimodal dataset consisting of upper-airway MRI scans of healthy French speakers[13]. This dataset is unique within the domain of speech MRI studies, as most datasets typically include only English speakers. The dataset comprises 2D mid-sagittal dynamic, and 3D vocal tract volumetric samples from 10 healthy native French speakers. Each 3D scan had a duration of 7 seconds and was acquired using a Siemens Prisma 3T scanner with a VIBE sequence (TR = 3.8, TE = 1.55, FOV = 22 × 22 cm², slice thickness = 1.2 mm, image size = 320 × 290 × 36 slices). In total, the dataset contains approximately 750 volumes, and we utilized only 53 3D volumes from all 10 subjects, considering the time required for manual annotation. Typical manual segmentation times range from 30 to 40 minutes per volume, depending on the complexity of the segmentation. A total of 21 unique French phonemes and 3 unique voiceless tasks were included. For training and validation, seven speakers were used, while three subjects were reserved for testing. Figure 1 illustrates examples from corresponding pre-training, training, and testing datasets along with their corresponding segmentations. We have aimed to choose training and testing data from all these volumes which provide different speech tasks. Adding a wider range of data provides more robust features to the network. Tables 1 A & B show the types of voiced and voiceless tasks we have chosen.

Prior to training and testing, different preprocessing methods were applied to CT and MRI volumes, respectively. For CT volumes, the voxel intensities were first clamped to the range of -1000 to 1000 of their original values. Subsequently, they were scaled to a range of 0 to 255 to represent pixel intensities. The segmentation maps corresponding to these CT volumes were binarized, with 0 representing the background and 1 representing the lung segmentation label. Similarly, for the MRI dataset, analogous preprocessing steps were undertaken. However, in addition to the aforementioned steps, gradient anisotropic diffusion (GAD)[43] was applied to reduce noise in the MRI data. Following this, voxel intensities were clamped to the range of 0 to 255. Furthermore, a cropping of 70 percent was applied specifically for MRI datasets to focus on the upper airway. Finally, both CT and MRI datasets were sampled to a size of 256 x 256 x 32 voxels. For deep learning models, maintaining fixed input dimensions across training and testing datasets is often necessary, requiring resampling or cropping to match the selected voxel size. Additionally, image sizes are commonly chosen as powers of 2, as most convolutional operations are optimized for such dimensions, helping prevent processing inefficiencies within the neural network. The original volumes had dimensions of 320 × 290 × 36 slices. After applying 70% cropping in the sagittal plane to focus on the upper airway, the images were resampled to a final size of 256 × 256 × 32. This selection balances resolution retention, computational efficiency, and memory constraints. Specifically, we mildly downsampled in the lateral direction while upsampling in the superior-inferior direction, ensuring a trade-off between preserving anatomical details and optimizing deep learning performance. We performed preprocessing steps in the Slicer environment equipped with Simple ITK libraries[44].

**Data Annotation:** The open-source volumetric French speaker MRI datasets lacked manually annotated airway segmentations which are needed to train a supervised deep learning segmentation algorithm. We selected a cohort of 53 volumes spanning a range of speech tasks, including 21 unique French phonemes, and 3 unique voiceless tasks. These were manually segmented in a two-stage process by an annotator team with expertise in voice-science. First, the volumes were distributed to two graduate students in voice science who performed the initial draft segmentations. Next, the draft segmentations were further reviewed slice by slice, and if any mistakes or missing regions were found, they were re-segmented by an expert vocologist with more than 20 years of experience in voice anatomy and singing voice pedagogy (co-author: D. Meyer). All processing was done in the 3D Slicer environment. Subsequently, the segmentations and the original MRI volumes underwent conversion into NRRD format. For training, we have used a subset of 45 volumes across 7 subjects from these 53 volumes.

Assessing the performance of an algorithm using annotated label maps from a single human expert as a reference poses challenges in evaluation due to potential bias introduced by the human. To mitigate this challenge, we employed the Simultaneous Truth and Performance Level Estimation (STAPLE) algorithm [45]. This method treats each annotator's segmentation as a noisy observation of an unknown true segmentation and iteratively estimates both the sensitivity and specificity of each annotator. Using an expectation-maximization approach, STAPLE computes a probabilistic estimate of the true segmentation by maximizing the likelihood of the observed annotations. In creating the test set, we selected eight volumes from three subjects (non-overlapping with the training set) and collected manual segmentations from three different annotators. The first two annotators were graduate students with expertise in image processing and biomedical engineering, and segmentations from a third annotator were those provided by the voice-science team as described above. These segmentations were used in the STAPLE algorithm to generate a consensus segmentation. The input segmentations from the various manual annotators and their corresponding STAPLE output are illustrated in Figure 2.

The collection of French speaker sounds utilized from the French speaker database for constructing the testing and the training datasets in this study are detailed in tables 1A and 1B. We selected a diverse set of phonemes across all the speakers. We use the same notations and format as reported in the original work, but explicitly detail the subject ids to highlight the subjects used in training are distinct from those used in testing.

***Data Augmentation:*** To augment the training dataset and increase its diversity, three types of data augmentations are applied during the creation of the training dataset: 1) noise addition:

Gaussian noise ranging from standard deviation of 0 to 0.01 were added to all 3D volumes; 2) flipping: volumes were flipped by 180 degrees, enhancing the variability of the dataset; 3) rotation: random rotation within the range of -10 to 10 degrees was applied to the volumes, further augmenting the dataset. These augmentation techniques contribute to the robustness and generalization ability of the neural network model by introducing variations in the training data.

***Evaluation & Metrics:*** The model's performance was evaluated against three subjects, which were used as testing sets. Instead of directly using one manual segmentation, we employed the STAPLE method, to generate the reference segmentation. Eight volumes across different speech postures were selected. We used the below quantitative metrics to evaluate the network performance.

*Dice coefficient (DC)[46]:* A statistic used to measure the similarity between two sets. In image segmentation, it evaluates the overlap between two segmented regions: a predicted segmentation (e.g., from an algorithm) and a ground truth segmentation (e.g., a manual segmentation by an expert). It ranges from 0 to 1, where 0 indicates no overlap and 1 signifies perfect agreement. The calculation involves twice the area of overlap divided by the total number of pixels in both images.

$$Dice(A, B) = \frac{(2|A \cap B|)}{(|A| + |B|)}$$

*Structural similarity Index measure (SSIM)[47]:* This measure tries to capture how a human would perceive the differences between two images, taking into account the structure of the images. It's based on the idea that the human visual system is highly sensitive to structural information. SSIM considers three factors: luminance, contrast, and structure. Two images might have the same average brightness and contrast, but if the *patterns* within the images are different, SSIM will be lower. The SSIM index for 3D volumes typically ranges from -1 to 1, where a score of 1 signifies perfect similarity between the volumes, 0 indicates no similarity, and -1 denotes complete dissimilarity.

$$SSIM(x, y) = \frac{\left((2\mu_x\mu_y + c_1)(2\sigma_{xy} + c_2)\right)}{\left((\mu_x^2 + \mu_y^2 + c_1)(\sigma_x^2 + \sigma_y^2 + c_2)\right)}$$

Where μx and μy are the mean intensities of images x and y, respectively. σxy is the covariance between x and y. σx² and σy² are the variances of x and y, respectively. c1 and c2 are constants to stabilize the division.

*Hausdorff distance (HD)[48]:* The Hausdorff distance is a metric used to measure how similar two sets of points are. Informally, it measures the greatest distance between a point in one set and the closest point in the other set, and vice-versa. It's particularly useful when comparing shapes or segmentations, as it considers the overall "closeness" of the sets, rather than just point-to-point correspondences. A smaller Hausdorff distance indicates greater similarity between the sets. Formally, given two sets $X$ and $Y$ in metric space, the Hausdorff distance $d_H(X,Y)$

$$d_H(X,Y) = max\{sup_{x \in X} inf_{y \in Y} d(x,y), sup_{y \in Y} inf_{x \in X} d(x,y)\}$$

This equation can be interpreted as follows: For each point in X, find the closest point in Y and calculate the distance and take the supremum (least upper bound) of all these distances. Repeat with X and Y swapped. The Hausdorff distance is the maximum of these two values.

### *Network architectures:*

We evaluated four variations of the U-Net architecture as distinct networks. The original U-Net architecture[49], introduced by Ronneberger et al. (2015), has achieved widespread recognition and adoption within the field of biomedical image segmentation. Its innovative design, characterized by a U-shaped network of convolutional layers, effectively captures both high-level contextual information and fine-grained spatial details. This ability to integrate multi-scale information has proven particularly advantageous in tasks requiring precise delineation of anatomical structures or pathological regions, as demonstrated by its successful application in various segmentation challenges[50–52].

*A)2D slice-by-slice U-Net[53]:* The 2D slice-by-slice U-Net is a variant of the U-Net architecture designed specifically for 2D medical image segmentation within 3D volumes, such as CT or MRI scans[31,38]. In this approach, the 3D volume is processed one slice at a time, treating each 2D slice as an independent input to the network. The network architecture follows an encoder-decoder structure with skip connections. The encoder gradually reduces the spatial dimensions of the input slice while extracting features through convolutional and pooling layers. On the other hand, the decoder up samples these features back to the original input size using transposed convolutions, while also incorporating skip connections from the encoder to preserve spatial information. This slice-by-slice processing allows the network to focus on the details within each 2D slice, which is advantageous for certain medical images where important features are primarily in the plane of the slice. However, it may overlook contextual information provided by neighboring slices, which could be relevant for accurate segmentation in some cases.

*B) 3D U-Net[50]:* In medical imaging, the 3D U-Net (Figure 3b) extends the original 2D U-Net to process 3D volumes. It consists of an encoder (downsampling path) and a decoder (upsampling path). The encoder progressively reduces the spatial dimensions (height, width, depth) of the input image while increasing the number of feature channels (feature maps) to capture high-level contextual information. The decoder then reconstructs the segmentation by progressively restoring spatial resolution while reducing the number of feature channels. This final stage produces the segmentation predictions. We employed the Dice coefficient as the common loss function for all U-Net architectures.

*C) 3D U-Net with transfer learning[54]:* In this approach, the previously mentioned 3D U-Net is first pre-trained using the OSIC lung dataset for lung segmentation. Subsequently, the top layers of this network are frozen, and the bottom layers are retrained using a French speaker dataset. For transfer learning, only 20 samples from the French speaker dataset were utilized. From our previous research[54], we found that going below 20 samples for re-training can significantly introduce non-anatomical segmentations. Pre-training enables the network to grasp low-level features specific to biomedical images, such as pixel intensity differences. Meanwhile, more abstract features are gleaned from the training dataset (French speaker dataset). This method allows for training the network with a small number of samples from the target dataset. This approach has been successfully implemented in prior work for segmenting vocal tract in 2D mid-sagittal dynamic MRI and has demonstrated the ability to achieve segmentation quality comparable to that of larger datasets[32].

*D) 3D U-NetR[51]:* A transformer-based variant of the U-Net architecture. Standard convolutional neural networks, while effective for local feature extraction, are limited in their ability to capture long-range dependencies within an image. The U-NetR addresses this limitation by incorporating a transformer encoder. Inspired by the success of transformers in natural language processing, the U-NetR treats the 3D image volume as a sequence, allowing the transformer to learn relationships between distant voxels and capture global contextual information. This transformer encoder is integrated within the established U-shaped U-Net framework. The decoder, through skip connections at multiple resolutions, combines these global representations with finer-grained local features from earlier convolutional layers, producing the final, detailed segmentation output. This architecture therefore leverages both the local feature extraction capabilities of convolutions and the global contextual understanding afforded by the transformer, resulting in improved segmentation accuracy. (see Fig. 3d).

All the given network architectures require an input size of 256 x 256 x 32 voxels and corresponding segmentations in the same size.

*Implementation and hyper-parameter tuning:* Hyperparameter tuning is essential for optimizing UNet model's performance. It directly impacts crucial outcome indicators like segmentation accuracy (e.g., Dice score), generalization to unseen data, and the speed and stability of the training process. By carefully adjusting hyperparameters such as learning rate, batch size, and model architecture, we can significantly improve the model's ability to accurately segment target structures, reduce overfitting, and ensure efficient resource utilization, ultimately leading to more robust and reliable results. In this study, all the networks are developed using MONAI[55] on an NVIDIA A30 GPU. We used grid-based search to tune all the hyperparameters for each of these networks. For the transfer learning network, the number of layers to be frozen was also considered as a hyperparameter. The hyperparameters evaluated in this study include the following: 1) Number of epochs: [60, 100, 200, 700, 1000, 1500, 50000], 2) Steps per epoch: [50, 100, 150, 200], 3) Learning rate: [1e-4, 3e-4, 3e-5, 1e-5], 4) Dropout rate: [0, 0.1, 0.5]. For the transfer learning network, an additional hyperparameter the number of layers to be frozen was also explored, with values ranging from [3, 5, 10, 15, 20, 25, 30, 35]. The optimal values for each hyperparameter were determined iteratively by running multiple configurations across the specified ranges for each network.

**RESULTS:**

Figure 4 displays the mid-sagittal slices of five representative test volumes, with each column corresponding to a different sustained sound. The four sounds are labeled using phonetic symbols: /f/, /l/ /k/, /a/. The label "UP" represents an unvoiced (silent) task, when the tongue is in contact with the upper teeth. We presume this is the vocal posture of /n/ but that it was an unvoiced posture as in the original paper[13]. The focus on the mid-sagittal plane in Figure 4 provides a clear view of the vocal tract anatomy and makes it easier to identify areas where the different models struggle to segment. Directly comparing the model outputs to the reference segmentation allows for a qualitative assessment of the performance of each model. For example, disjoint masks, which are a type of segmentation error where the model identifies small, isolated areas of airspace that should not be included in the vocal tract segmentation, are visible in the third, fourth, and fifth columns of the third, fourth, and fifth rows. These errors are most apparent in the 3D U-Net, 3D U-Net with transfer learning, and 3D U-NetR models. The /k/ sound, displayed in the fourth column, appears to be particularly challenging, as all four models exhibit segmentation errors in this column. We also note that in the "UP" posture (voiceless task, where the tongue tip hits the surface of the front teeth), the airspace behind the velum is missed by all the models. This may be attributed to the low number of voiceless tasks in the training data (only 3 of the 45 volumes), and the models may have learnt patterns present from the sustained voicing tasks dominant in the training data, where the velopharyngeal port is closed.

Figure 5 provides a three-dimensional visualization of the vocal tract airspace volume for the same five test volumes depicted in Figure 4. This figure allows for a more holistic assessment of the segmentation results, taking into account the overall shape and curvature of the vocal tract airspace. The 3D U-NetR model generated numerous non-anatomical segments that deviate significantly from the reference segmentation. The segmentations produced by the 2D slice-by-slice U-Net model often had rough edges and, in some cases, included large, non-anatomical segments. One notable example is the segmentation for the /a/ sound (last column of Figure 5), where the 2D slice-by-slice U-Net model incorrectly includes a large portion of the area near the epi-glottis in the segmentation. Visually, the 3D U-Net and the 3D U-Net with transfer learning models produced very similar segmentations. However, the 3D U-Net model required 45 training volumes, while the 3D U-Net with transfer learning model only needed 20 training volumes to achieve comparable results. This finding suggests that transfer learning could be a valuable strategy for vocal tract segmentation with limited annotated samples.

Table 2 displays the quantitative segmentation results for the four models across eight volumes. The metrics used to evaluate performance include Dice Coefficient, Hausdorff distance (HD distance), and structural similarity index measure (SSIM Index). Individual volume performance as well as the average and standard deviation for each metric are included. The 2D slice-by-slice U-Net model consistently achieved lower Dice values than the other three methods, with an average score of 0.823 ± 0.05. The 3D U-Net and the transfer learning 3D U-Net models both achieved the highest average Dice scores (0.896 ± 0.05 and 0.896 ± 0.04, respectively). The results of SSIM show similar trend as the Dice scores. The table highlights that volumes 8 (/ʃ/) and 3 ("UP") consistently have the lowest Dice scores across all models, with volume 8 showing the most significant discrepancy. The 2D slice-by-slice U-Net method achieved a relatively consistent performance for HD distance with an average of 11.3 ± 5.4, lower than the

3D U-Net (14 ± 28) and the transfer learning 3D U-Net (15 ± 24.9). The 3D U-Net transfer learning model showed the best HD distance with an average of (3.95 ± 5.2). Despite the higher average Dice scores, the 3D U-Net and 3D U-NetR methods showed greater variability in HD distance (note considerably higher standard deviations), which might indicate inconsistencies in the model's ability to delineate boundaries.

A qualitative assessment of the network results was performed by the co-author D. Meyer, an expert vocologist. In this context, "acceptable" refers to segmentations that align well with anatomical structures, without introducing non-anatomical errors. The voice expert assessed acceptability by examining both mid-sagittal slices and volumetric surface renders, identifying any segmentation artifacts or deviations from expected anatomy. Based on this evaluation, the segmentation of the oral cavity airspace was deemed acceptable for all network results except the 2D slice-by-slice U-Net. Similarly, the oropharyngeal and laryngopharyngeal segmentations were considered acceptable when compared to the reference segmentations. All models frequently mis-segmented islands of airspace in either the tongue, lower incisors (anterior mandible), or in the palatine process of the maxilla (commonly known as the hard palate). Correctly imaging the teeth and maxilla is a known challenge for MRI. These structures are low in hydrogen, resulting in signal voids in MRI. This issue is particularly relevant for segmenting the oral airspace, which is in close proximity to the teeth and maxilla. Glottal and supraglottal structures often contained mis-segmentations as well. Inaccuracies in this region were not unexpected due to the difficulty in isolating glottal landmarks in manually segmented MRI volumes. 3D U-Net transformer showed frequent non-anatomical segmentations. 3D U-Net and transfer-learning 3D U-Net segmentations were judged to most closely resemble the reference segmentations in the glottal and supraglottic airspace.

**DISCUSSION:**

This study provided a rich manually annotated public database of vocal tract segmentations from a cohort of 53 volumes across 10 speakers producing French phonemes. The study compared neural network architectures for segmenting the upper airway from 3D MRI, focusing on convolutional networks, transformers, and transfer learning methods. Transfer learning proved particularly effective in addressing the challenges of training with small datasets, a common problem in specialized areas where extensive, open-source datasets are lacking. This approach is especially relevant for vocal tract segmentation as it is a relatively niche area of research and publicly available annotated datasets are scarce.

The variability in manual segmentation results highlighted the need for a consistent approach to creating ground truth segmentations. The STAPLE algorithm was employed to mitigate the potential bias introduced by individual annotators. This method enhances the stability of quantitative metrics by generating a probabilistic estimate of the true segmentation based on the input of segmentations from multiple human experts.

Comparisons of different U-Net architectures in medical imaging revealed that 3D convolutional networks consistently outperformed transformer-based U-Net and slice-by-slice U-Net architectures. This suggests that 3D convolutional networks are well-suited for medical imaging applications where datasets may be limited, as they can effectively learn from smaller amounts of training data. Both the 3D U-Net and the 3D U-Net with transfer learning achieved high average Dice coefficients (0.896 ± 0.05 and 0.896 ± 0.04, respectively). Importantly, the transfer learning approach achieved performance comparable to the standard 3D U-Net while using less than half the training data. This underscores the potential of transfer learning to optimize resource utilization in vocal tract segmentation tasks. The 2D slice-by-slice U-Net consistently produced lower Dice coefficients than other methods (0.823 ± 0.05 on average). While this architecture can effectively segment individual 2D slices, it is limited in its ability to leverage contextual information from neighboring slices, which could be contributing to its lower performance. The 3D Transformer U-Net is designed to capture long-range spatial dependencies, however, in this work, it generated non-anatomical segmentations that deviated significantly from the ground truth. This suggests that the transformer architecture may not be as well-suited for this specific segmentation task, potentially due to the complexity of the vocal tract anatomy and limited training data.

A qualitative assessment by a voice expert revealed that all methods appeared to delineate the oral airspace, except the 2D slice by slice U-Net. All models frequently incorrectly segmented islands of airspace in the tongue or hard palate, as well as the supraglottic space. These errors are likely attributed to the difficulty in clearly identifying anatomical landmarks in MRI, particularly in the glottal and supraglottic regions. The presence of teeth in the vocal tract results in signal voids and pose a challenge for MRI analysis, especially for segmentation of the oral airspace which is in close proximity to the teeth.

We used the STAPLE method to integrate segmentations from three annotators to estimate a probabilistic ground truth in the testing set. However, STAPLE can be sensitive to systematic errors; if annotators make consistent mistakes, the algorithm may reinforce rather than correct them, leading to deviations from the true segmentation. While STAPLE weights annotators by reliability, frequent biases can still affect the final result. To mitigate this, we selected a diverse set of annotators with expertise in upper-airway anatomy: two biomedical engineering graduate students with experience in image processing and upper-airway anatomy, and a vocologist with 20 years of voice science pedagogy experience.

Our study has a few noteworthy limitations. First, the small size of the training dataset is a potential limitation.  While data augmentation techniques were employed to increase the diversity of the training data, the relatively small number of annotated volumes (45 for training) may have limited the models' ability to generalize to unseen data, which is particularly challenging in vocal tract segmentation. A larger and more diverse training dataset encompassing a wider range of speakers, phonetic contexts, and vocal tract postures would likely improve the accuracy and robustness of the models. Secondly, we utilized a specific MRI acquisition protocol for the French speaker dataset. The performance of the trained models may vary when applied to data acquired using different MRI scanners, sequences, or parameters. While the transfer learning model may be a useful approach to address this limitation, further research is necessary to assess the generalizability across diverse MRI acquisition protocols. We also note noteworthy challenges by all the models for specific tasks, such as segmentation of vocal tract while producing the /k/ sound, and during un-voiced /n/ (i.e, the tongue touching the upper teeth in the "UP" posture). These highlight the complexity of vocal tract anatomy and the difficulty in accurately segmenting highly variable and nuanced articulatory postures. Additionally, the models struggled with segmenting "islands of airspace," which appear as small, isolated pockets of air trapped within the vocal tract. These islands are mis-segmentations that occurred in areas such as the tongue, hard palate, and supraglottic space, and are likely due to the limitations of low-resolution MRI in differentiating airspace from surrounding tissues. We also note the issue of signal voids in MRI caused by the presence of teeth, which can hinder the accurate segmentation of the oral airspace. Teeth, being low in hydrogen content, have a short T2 time constant, and appear dark in conventional gradient echo MR imaging, making it difficult to distinguish them from the surrounding air. This challenge can lead to errors in defining the boundaries of the oral cavity, particularly near the teeth. Emerging techniques such as zero-echo time MRI[56] which provide teeth and bone visualization has promise to address this limitation.

**CONCLUSION:**

This study investigated the efficacy of deep learning algorithms for automatic vocal tract segmentation from 3D MRI. A new open-source database of 53 manually annotated 3D vocal tract volumes from 10 French speakers was created, adding diversity to the field of voice and speech science, which is predominantly focused on English speakers. 3D convolutional neural networks, specifically the 3D U-Net and the 3D U-Net with transfer learning, demonstrated superior performance compared to the 2D slice-by-slice U-Net and the 3D transformer U-Net models. The 3D U-Net with transfer learning achieved high accuracy while using less than half of the training data required by the 3D U-Net, highlighting its potential utility in vocal tract segmentation. The STAPLE algorithm, employed to generate a probabilistic estimate of the true segmentation from multiple annotators, enhanced the reliability of the evaluation process. Qualitative analysis by a voice expert revealed challenges in segmenting islands of airspace in the tongue or hard palate, and the supraglottic space. These difficulties can be attributed to the limitations of MRI in imaging structures with low hydrogen content, such as teeth, and the inherent complexities of accurately defining glottal and supraglottic landmarks.

**DATA AND CODE AVAILABILITY:**

The manually annotated volumes, along with their corresponding segmentations, are provided for download at Figshare (https://figshare.com/s/cb050b61c0189605feda). Additionally, to ensure the robustness of our method, the test set includes STAPLE segmentations derived from three distinct manual segmentations, further enriching the dataset. In addition to the dataset the python code used to run the individual networks can be downloaded from https://github.com/eksubin/Comparative-Study-of-3D-2D-Transfer-Learning-UNet.

**ACKNOWLEDGEMENT:** This work was supported by National Institute of Health (NIH) under grant NIH NHLBI: R01 HL173483.

**REFERENCES:**

1. Katz, W. F., Mehta, S., Wood, M., & Wang, J. (2017). Using electromagnetic articulography with a tongue lateral sensor to discriminate manner of articulation. *The Journal of the Acoustical Society of America*, *141*(1), EL57. https://doi.org/10.1121/1.4973907

2. Döllinger, M., Kunduk, M., Kaltenbacher, M., Vondenhoff, S., Ziethe, A., Eysholdt, U., & Bohr, C. (2012). Analysis of Vocal Fold Function From Acoustic Data Simultaneously Recorded With High-Speed Endoscopy. *Journal of Voice*, *26*(6), 726–733. https://doi.org/10.1016/j.jvoice.2012.02.001

3. Badin, P. (1991). Fricative consonants: acoustic and X-ray measurements. *Journal of Phonetics*, *19*(3–4), 397–408. https://doi.org/10.1016/S0095-4470(19)30331-6

4. Kabaliuk, N., Nejati, A., Loch, C., Schwass, D., Cater, J. E., & Jermy, M. C. (2017). Strategies for Segmenting the Upper Airway in Cone-Beam Computed Tomography (CBCT) Data. *Open Journal of Medical Imaging*, *07*(04), 196–219. https://doi.org/10.4236/ojmi.2017.74019

5. Fabre, D., Hueber, T., Girin, L., Alameda-Pineda, X., & Badin, P. (2017). Automatic animation of an articulatory tongue model from ultrasound images of the vocal tract. *Speech Communication*, *93*, 63–75. https://doi.org/10.1016/j.specom.2017.08.002

6. Bresch, E., Kim, Y. C., Nayak, K., Byrd, D., & Narayanan, S. (2008). Seeing speech: Capturing vocal tract shaping using real-time magnetic resonance imaging. *IEEE Signal Processing Magazine*, *25*(3). https://doi.org/10.1109/MSP.2008.918034

7. Story, B. H., Titze, I. R., & Hoffman, E. A. (1998). Vocal tract area functions for an adult female speaker based on volumetric imaging. *The Journal of the Acoustical Society of America*, *104*(1), 471–487. https://doi.org/10.1121/1.423298

8. Lingala, S. G., Sutton, B. P., Miquel, M. E., & Nayak, K. S. (2016). Recommendations for real-time speech MRI. *Journal of Magnetic Resonance Imaging*, *43*(1), 28–44. https://doi.org/10.1002/jmri.24997

9. Bresch, E., Yoon-Chul Kim, Nayak, K., Byrd, D., & Narayanan, S. (2008). Seeing speech: Capturing vocal tract shaping using real-time magnetic resonance imaging [Exploratory DSP]. *IEEE Signal Processing Magazine*, *25*(3), 123–132. https://doi.org/10.1109/MSP.2008.918034

10. Lingala, S. G., Toutios, A., Toger, J., Lim, Y., Zhu, Y., Kim, Y. C., Vaz, C., Narayanan, S., & Nayak, K. (2016). State-of-the-art MRI protocol for comprehensive assessment of vocal tract structure and function. *Proceedings of the Annual Conference of the International Speech Communication Association, INTERSPEECH*. https://doi.org/10.21437/Interspeech.2016-559

11. Lim, Y., Toutios, A., Bliesener, Y., Tian, Y., Lingala, S. G., Vaz, C., Sorensen, T., Oh, M., Harper, S., Chen, W., Lee, Y., Töger, J., Monteserin, M. L., Smith, C., Godinez, B., Goldstein, L., Byrd, D., Nayak, K. S., & Narayanan, S. S. (2021). A multispeaker dataset of raw and reconstructed speech production real-time MRI video and 3D volumetric images. *Scientific Data 2021 8:1*, *8*(1), 1–14. https://doi.org/10.1038/s41597-021-00976-x

12. Burdumy, M., Traser, L., Burk, F., Richter, B., Echternach, M., Korvink, J. G., Hennig, J., & Zaitsev, M. (2017). One-second MRI of a three-dimensional vocal tract to measure dynamic articulator modifications. *Journal of Magnetic Resonance Imaging : JMRI*, *46*(1), 94–101. https://doi.org/10.1002/JMRI.25561

13. Isaieva, K., Laprie, Y., Leclère, J., Douros, I. K., Felblinger, J., & Vuissoz, P. A. (2021). Multimodal dataset of real-time 2D and static 3D MRI of healthy French speakers. *Scientific Data 2021 8:1*, *8*(1), 1–9. https://doi.org/10.1038/s41597-021-01041-3

14. Belyk, M., Waters, S., Kanber, E., Miquel, M. E., & McGettigan, C. (2022). Individual differences in vocal size exaggeration. *Scientific Reports 2022 12:1*, *12*(1), 1–12. https://doi.org/10.1038/s41598-022-05170-6

15. Story, B. H., Titze, I. R., & Hoffman, E. A. (1998). Vocal tract area functions from magnetic resonance imaging. *The Journal of the Acoustical Society of America*, *100*(1), 537. https://doi.org/10.1121/1.415960

16. Story, B. H., Titze, I. R., & Hoffman, E. A. (2001). The relationship of vocal tract shape to three voice qualities. *The Journal of the Acoustical Society of America*, *109*(4), 1651. https://doi.org/10.1121/1.1352085

17. Story, B. H. (2013). Phrase-level speech simulation with an airway modulation model of speech production. *Computer Speech & Language*, *27*(4), 989–1010. https://doi.org/10.1016/J.CSL.2012.10.005

18. Rusho, R. Z., Meyer, D., Jacob, M., Story, B., & Lingala, S. G. (2023). *Synthesizing speech through a tube talker model informed by dynamic MRI-derived vocal tract area functions*. Proc Intl Soc Mag Reson Med.

19. Birkholz, P., Kürbis, S., Stone, S., Häsner, P., Blandin, R., & Fleischer, M. (2020). Printable 3D vocal tract shapes from MRI data and their acoustic and aerodynamic properties. *Scientific Data 2020 7:1*, *7*(1), 1–16. https://doi.org/10.1038/s41597-020-00597-w

20. Feng, M., & Howard, D. M. (2023). The Dynamic Effect of the Valleculae on Singing Voice – An Exploratory Study Using 3D Printed Vocal Tracts. *Journal of Voice*, *37*(2), 178–186. https://doi.org/10.1016/J.JVOICE.2020.12.012

21. Howard, D. M. (2018). The Vocal Tract Organ: A New Musical Instrument Using 3-D Printed Vocal Tracts. *Journal of Voice*, *32*(6), 660–667. https://doi.org/10.1016/J.JVOICE.2017.09.014

22. Javed, A., Kim, Y. C., Khoo, M. C. K., Ward, S. L. D., & Nayak, K. S. (2016). Dynamic 3-D MR visualization and detection of upper airway obstruction during sleep using region-growing segmentation. *IEEE Transactions on Biomedical Engineering*. https://doi.org/10.1109/TBME.2015.2462750

23. Skordilis, Z. I., Toutios, A., Toger, J., & Narayanan, S. (2017). Estimation of vocal tract area function from volumetric Magnetic Resonance Imaging. *ICASSP, IEEE International Conference on Acoustics, Speech and Signal Processing - Proceedings*. https://doi.org/10.1109/ICASSP.2017.7952291

24. Skordilis, Z. I., Ramanarayanan, V., Goldstein, L., & Narayanan, S. S. (2015). Experimental assessment of the tongue incompressibility hypothesis during speech production. *Proceedings of the Annual Conference of the International Speech Communication Association, INTERSPEECH*.

25. Labrunie, M., Badin, P., Voit, D., Joseph, A. A., Frahm, J., Lamalle, L., Vilain, C., & Boë, L. J. (2018). Automatic segmentation of speech articulators from real-time midsagittal MRI based on supervised learning. *Speech Communication*, *99*, 27–46. https://doi.org/10.1016/J.SPECOM.2018.02.004

26. Silva, S., & Teixeira, A. (2015). Unsupervised segmentation of the vocal tract from real-time MRI sequences. *Computer Speech & Language*, *33*(1), 25–46. https://doi.org/10.1016/J.CSL.2014.12.003

27. Sampaio, R. D. A., & Jackowski, M. P. (2017). Vocal Tract Morphology Using Real-Time Magnetic Resonance Imaging. *Proceedings - 30th Conference on Graphics, Patterns and Images, SIBGRAPI 2017*, 359–366. https://doi.org/10.1109/SIBGRAPI.2017.54

28. Ramanarayanan, V., Tilsen, S., Proctor, M., Töger, J., Goldstein, L., Nayak, K. S., & Narayanan, S. (2018). Analysis of speech production real-time MRI. *Computer Speech & Language*, *52*, 1–22. https://doi.org/10.1016/J.CSL.2018.04.002

29. Bresch, E., & Narayanan, S. (2009). Region segmentation in the frequency domain applied to upper airway real-time magnetic resonance images. *IEEE Transactions on Medical Imaging*. https://doi.org/10.1109/TMI.2008.928920

30. Ruthven, M., Miquel, M. E., & King, A. P. (2021). Deep-learning-based segmentation of the vocal tract and articulators in real-time magnetic resonance images of speech. *Computer Methods and Programs in Biomedicine*. https://doi.org/10.1016/j.cmpb.2020.105814

31. Erattakulangara, S., & Lingala, S. G. (2020). Airway segmentation in speech MRI using the U-net architecture. *IEEE International Symposium on Biomedical Imaging (ISBI)*, to appear.

32. Erattakulangara, S., Kelat, K., Meyer, D., Priya, S., Lingala, S. G., & Carver, R. J. (2023). Automatic Multiple Articulator Segmentation in Dynamic Speech MRI Using a Protocol Adaptive Stacked Transfer Learning U-NET Model. *Bioengineering 2023, Vol. 10, Page 623*, *10*(5), 623. https://doi.org/10.3390/BIOENGINEERING10050623

33. Valliappan, C. A., Kumar, A., Mannem, R., Karthik, G. R., & Ghosh, P. K. (2019). An Improved Air Tissue Boundary Segmentation Technique for Real Time Magnetic Resonance Imaging Video Using Segnet. *ICASSP, IEEE International Conference on Acoustics, Speech and Signal Processing - Proceedings*, *2019-May*, 5921–5925. https://doi.org/10.1109/ICASSP.2019.8683153

34. Valliappan, C. A., Mannem, R., & Kumar Ghosh, P. (2018). Air-tissue boundary segmentation in real-time magnetic resonance imaging video using semantic segmentation with fully convolutional networks. *Proceedings of the Annual Conference of the International Speech Communication Association, INTERSPEECH*. https://doi.org/10.21437/Interspeech.2018-1939

35. Xie, L., Udupa, J. K., Tong, Y., Torigian, D. A., Huang, Z., Kogan, R. M., Wootton, D., Choy, K. R., Sin, S., Wagshul, M. E., & Arens, R. (2022). Automatic upper airway segmentation in static and dynamic MRI via anatomy-guided convolutional neural networks. *Medical Physics*, *49*(1), 324–342. https://doi.org/10.1002/MP.15345

36. Xie, L., Udupa, J. K., Tong, Y., Torigian, D. A., Huang, G., Kogan, R. M., Nathan, J. Ben, Wootton, D. M., Choy, K. R., Sin, S., Wagshul, M., & Arens, R. (2021). Automatic upper airway segmentation in static and dynamic MRI via deep convolutional neural networks. *Https://Doi.Org/10.1117/12.2581974*, *11600*, 131–136. https://doi.org/10.1117/12.2581974

37. Ma, D., Gulani, V., Seiberlich, N., Liu, K., Sunshine, J. L., Duerk, J. L., & Griswold, M. a. (2013). Magnetic resonance fingerprinting. *Nature*, *495*(7440), 187–192. https://doi.org/10.1038/nature11971

38. Bommineni, V. L., Erus, G., Doshi, J., Singh, A., Keenan, B. T., Schwab, R. J., Wiemken, A., & Davatzikos, C. (2023). Automatic Segmentation and Quantification of Upper Airway Anatomic Risk Factors for Obstructive Sleep Apnea on Unprocessed Magnetic Resonance Images. *Academic Radiology*, *30*(3), 421–430. https://doi.org/10.1016/J.ACRA.2022.04.023

39. Ruthven, M., Peplinski, A. M., Adams, D. M., King, A. P., & Miquel, M. E. (2023). Real-time speech MRI datasets with corresponding articulator ground-truth segmentations. *Scientific Data 2023 10:1*, *10*(1), 1–10. https://doi.org/10.1038/s41597-023-02766-z

40. Warfield, S. K., Zou, K. H., & Wells, W. M. (2004). Simultaneous truth and performance level estimation (STAPLE): An algorithm for the validation of image segmentation. *IEEE Transactions on Medical Imaging*, *23*(7), 903–921. https://doi.org/10.1109/TMI.2004.828354

41. *OSIC Pulmonary Fibrosis Progression | Kaggle*. (n.d.). Retrieved December 6, 2024, from https://www.kaggle.com/competitions/osic-pulmonary-fibrosis-progression

42. *CT Lung & Heart & Trachea segmentation*. (n.d.). Retrieved December 6, 2024, from https://www.kaggle.com/datasets/sandorkonya/ct-lung-heart-trachea-segmentation

43. Perona, P., & Malik, J. (1990). Scale-space and edge detection using anisotropic diffusion. *IEEE Transactions on Pattern Analysis and Machine Intelligence*, *12*(7), 629–639. https://doi.org/10.1109/34.56205

44. Pieper, S., Halle, M., & Kikinis, R. (n.d.). 3D Slicer. *2004 2nd IEEE International Symposium on Biomedical Imaging: Macro to Nano (IEEE Cat No. 04EX821)*, *2*, 632–635. https://doi.org/10.1109/ISBI.2004.1398617

45. Warfield, S. K., Zou, K. H., & Wells, W. M. (2004). Simultaneous truth and performance level estimation (STAPLE): An algorithm for the validation of image segmentation. *IEEE Transactions on Medical Imaging*, *23*(7), 903–921. https://doi.org/10.1109/TMI.2004.828354

46. Bertels, J., Eelbode, T., Berman, M., Vandermeulen, D., Maes, F., Bisschops, R., & Blaschko, M. (2019). *Optimizing the Dice Score and Jaccard Index for Medical Image Segmentation: Theory & Practice*. https://doi.org/10.1007/978-3-030-32245-8_11

47. Wang, Z., Bovik, A. C., Sheikh, H. R., & Simoncelli, E. P. (2004). Image Quality Assessment: From Error Visibility to Structural Similarity. *IEEE Transactions on Image Processing*, *13*(4), 600–612. https://doi.org/10.1109/TIP.2003.819861

48. Hausdorff, F. (1914). *Grundzüge der Mengenlehre*. Leipzig Viet.

49. Ronneberger, O., Fischer, P., & Brox, T. (2015). *U-Net: Convolutional Networks for Biomedical Image Segmentation* (pp. 234–241). https://doi.org/10.1007/978-3-319-24574-4_28

50. Çiçek, Ö., Abdulkadir, A., Lienkamp, S. S., Brox, T., & Ronneberger, O. (2016). *3D U-Net: Learning Dense Volumetric Segmentation from Sparse Annotation* (pp. 424–432). https://doi.org/10.1007/978-3-319-46723-8_49

51. Hatamizadeh, A., Tang, Y., Nath, V., Yang, D., Myronenko, A., Landman, B., Roth, H., & Xu, D. (2021). *UNETR: Transformers for 3D Medical Image Segmentation*. http://arxiv.org/abs/2103.10504

52. Bakas, S., Reyes, M., Jakab, A., Bauer, S., Rempfler, M., Crimi, A., Shinohara, R. T., Berger, C., Ha, S. M., Rozycki, M., Prastawa, M., Alberts, E., Lipkova, J., Freymann, J., Kirby, J., Bilello, M., Fathallah-Shaykh, H., Wiest, R., Kirschke, J., … Menze, B. (2018). *Identifying the Best Machine Learning Algorithms for Brain Tumor Segmentation, Progression Assessment, and Overall Survival Prediction in the BRATS Challenge*.

53. Xie, L., Udupa, J. K., Tong, Y., Torigian, D. A., Huang, Z., Kogan, R. M., Wootton, D., Choy, K. R., Sin, S., Wagshul, M. E., & Arens, R. (2022). Automatic upper airway segmentation in static and dynamic MRI via anatomy-guided convolutional neural networks. *Medical Physics*, *49*(1), 324–342. https://doi.org/10.1002/mp.15345

54. Erattakulangara, S., Kelat, K., Meyer, D., Priya, S., & Lingala, S. G. (2023). Automatic Multiple Articulator Segmentation in Dynamic Speech MRI Using a Protocol Adaptive Stacked Transfer Learning U-NET Model. *Bioengineering*, *10*(5). https://doi.org/10.3390/bioengineering10050623

55. Cardoso, M. J., Li, W., Brown, R., Ma, N., Kerfoot, E., Wang, Y., Murrey, B., Myronenko, A., Zhao, C., Yang, D., Nath, V., He, Y., Xu, Z., Hatamizadeh, A., Myronenko, A., Zhu, W., Liu, Y., Zheng, M., Tang, Y., … Feng, A. (2022). *MONAI: An open-source framework for deep learning in healthcare*.

56. Aydıngöz, Ü., Yıldız, A. E., & Ergen, F. B. (2022). Zero Echo Time Musculoskeletal MRI: Technique, Optimization, Applications, and Pitfalls. *Radiographics*, *42*(5), 1398–1414. https://doi.org/10.1148/RG.220029/ASSET/IMAGES/LARGE/RG.220029.FIG18.JPEG

**FIGURES**

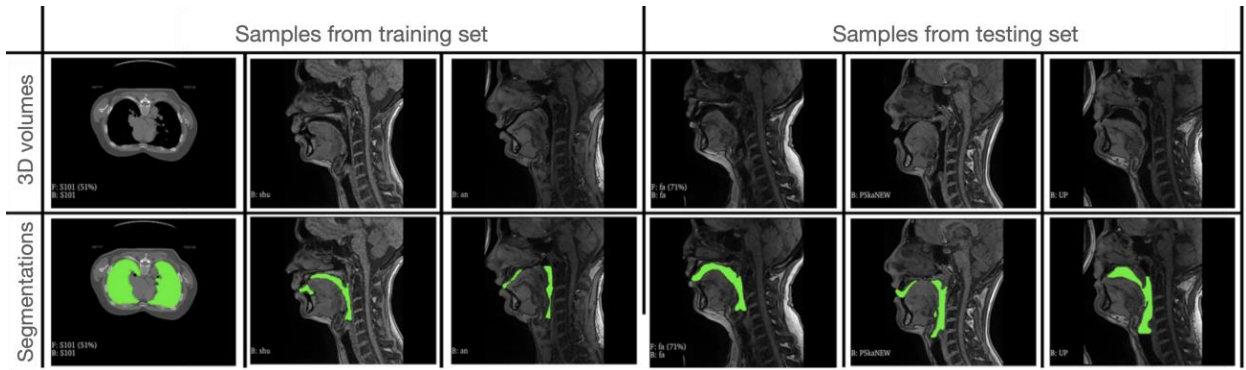

Figure 1: Sample images from the training and testing datasets. The first column displays a sample from the CT dataset used for pre-training the 3D U-Net transfer learning model. Columns 2 and 3 show samples from the training dataset used for all other neural networks. The color maps indicate the segmentation overlaid on the mid-sagittal section for MRI and the mid-axial cut of the CT dataset. Testing was conducted on three subjects not included in the training dataset. Sample vocal tract postures across speech postures from these subjects are shown in columns 4-6.

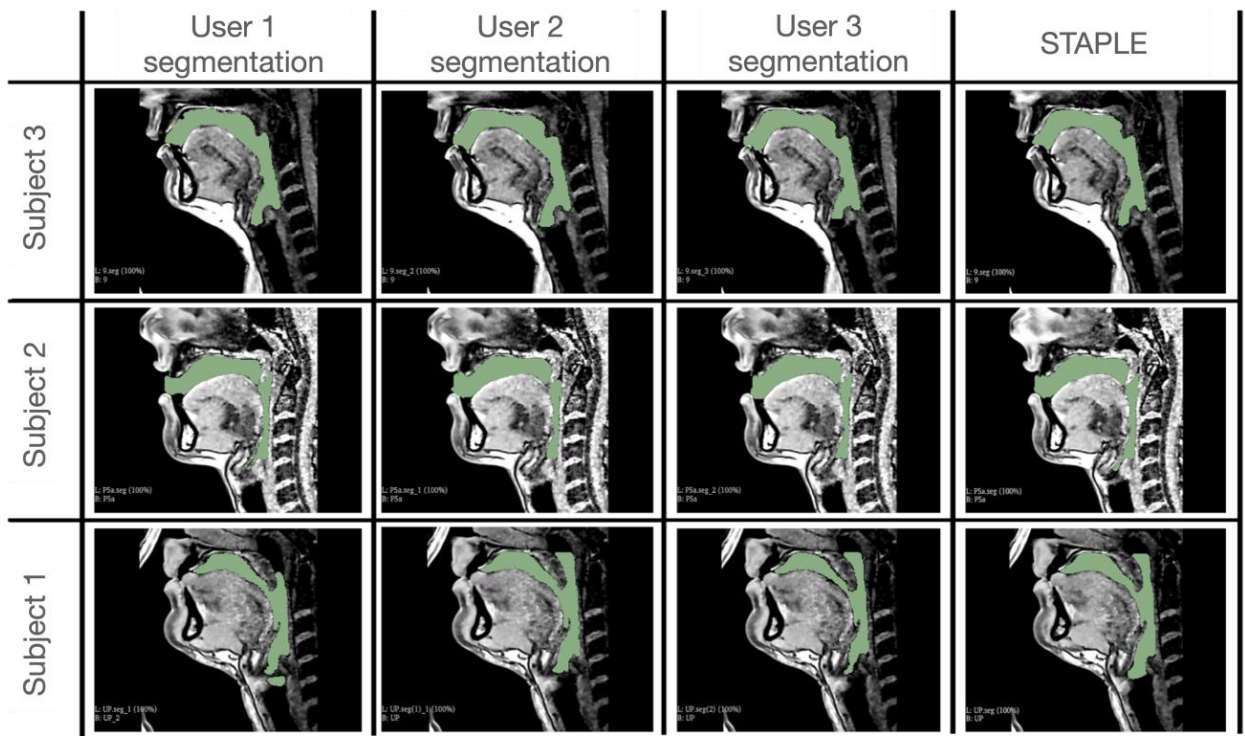

*Figure 2: Representation of the STAPLE method to create ground truth segmentations. Columns 1-3 show individual manual user segmentations respectively from three different manual users. Each row shows example vocal tract postures from three different subjects in the French speaker database. The final ground truth segmentation is generated using a probabilistic voting-based STAPLE algorithm by combining the three user segmentations to mitigate inter-user variability.*

| File name | Subject Number | Phoneme / position | In the context of phoneme | Word example |
|---|---|---|---|---|
| 9 | 4 | /oe/ | | peur |
| a | 5 | /a/ | | pas |
| UP | 6 | Tongue pushed against upper teeth | | |
| fa | 4 | /f/ | /a/ | fa |
| li | 4 | /l/ | /i/ | lit |
| kon | 6 | /k/ | /õ/ | con |
| nu | 5 | /n/ | /u/ | nous |
| sh2 | 4 | /ʃ/ | /ø/ | cheveu |

*Table 1A: Shows the testing dataset with different phoneme positions during sustained voicing. The phoneme contexts and word examples are also shown.*

| File name | Subject Numbers | Phoneme / position | In the context of phoneme | Word example |
|---|---|---|---|---|
| DOWN | 2 | Tongue pushed against lower teeth | | |
| CONTACT | 2 | Incisors in contact | | |
| UP | 2 | Tongue pushed against upper teeth | | |
| e | 7 | /e/ | | p (letter of the French alphabet) |
| a | 10 | /a/ | | pas |
| o | 1 | /o/ | | peau |
| y | 2 | /y/ | | pu |
| an | 2 | /ã/ | | pan |
| ri | 8 | /ʁ/ | /i/ | riz |
| ren | 1 | /ʁ/ | /ɛ̃/ | rein |
| pi | 2, 7, 10 | /p/ | /i/ | pis |
| pa | 1, 2, 7, 10 | /p/ | /a/ | pas |
| pu | 1, 7 | /p/ | /u/ | pou |
| py | 1, 2, 7 | /p/ | /y/ | pu |
| tu | 1 | /t/ | /u/ | tout |
| ki | 8 | /k/ | /i/ | qui |
| ka | 2, 7 | /k/ | /a/ | cadeau |
| ko | 7, 8 | /k/ | /o/ | colonie |
| ku | 7, 8 | /k/ | /u/ | cou |
| k2 | 7 | /k/ | /ø/ | queue |
| shE | 8 | /ʃ/ | /ɛ/ | chaise |
| sa | 1 | /s/ | /a/ | sa |
| so | 7 | /s/ | /o/ | sceau |
| s2 | 8 | /s/ | /ø/ | ceux |
| fi | 1,10 | /f/ | /i/ | fit |
| mi | 7, 8, 10 | /m/ | /i/ | mie |
| ma | 1, 7 | /m/ | /a/ | ma |
| mon | 8 | /m/ | /õ/ | mon |
| wa | 2, 7, 10 | /w/ | /a/ | voiture |

*Table 1B: Shows the training dataset with different phoneme positions during sustained voicing. The phoneme contexts and word examples are also shown.*

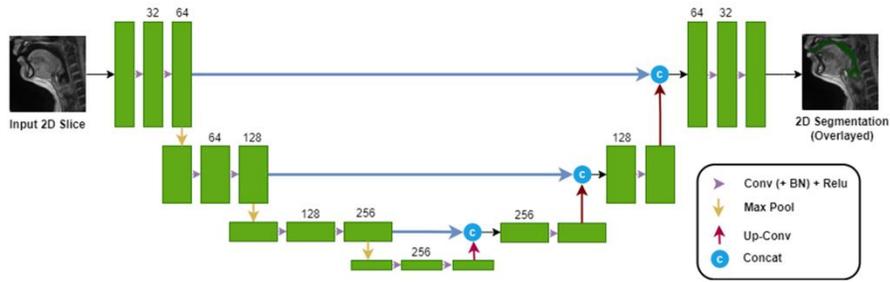

**3(a): 2D Slice-by-slice Unet**

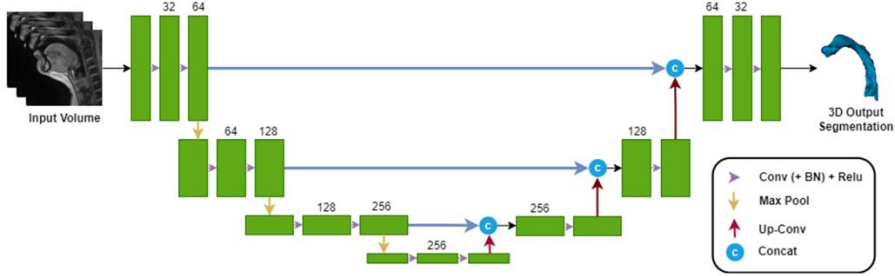

**3(b): 3D Unet**

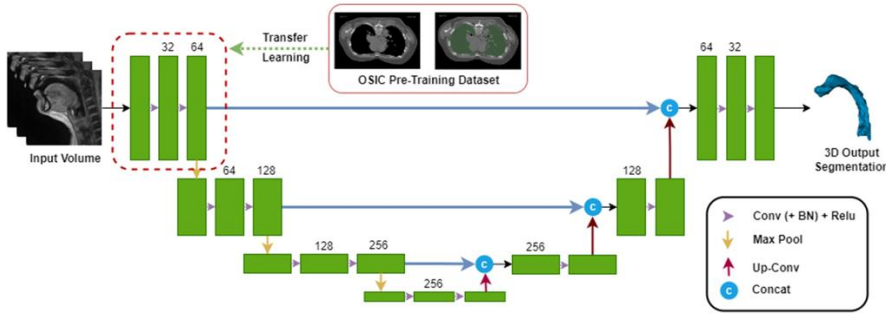

**3(c): 3D Unet with transfer learning**

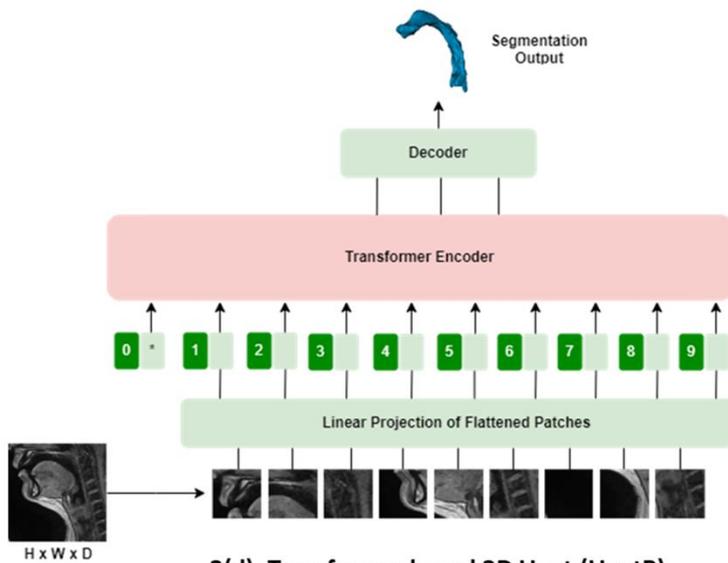

**3(d): Transformer based 3D Unet (UnetR)**

*Figure 3: showcases the diverse neural network architectures employed in this study. Subfigure 3(a) illustrates the 2D slice-by-slice U-Net utilized for segmentation generation from 2D slices, featuring 2D convolution operations. Subfigure 3(b) displays the 3D U-Net architecture incorporating 3D convolutions, operating on 3D volumes as input. Subfigure 3(c) presents a customized version of the standard 3D U-Net, where initial layers are pre-trained with the OSIC pulmonary dataset and kept frozen to preserve information, while subsequent layers are trained on a dataset smaller than the size of the original training set. Lastly, subfigure 3(d) introduces the transformer-based U-Net, which processes input volumes by dividing them into multiple patches to facilitate learning and segmentation map generation for the upper airway.*

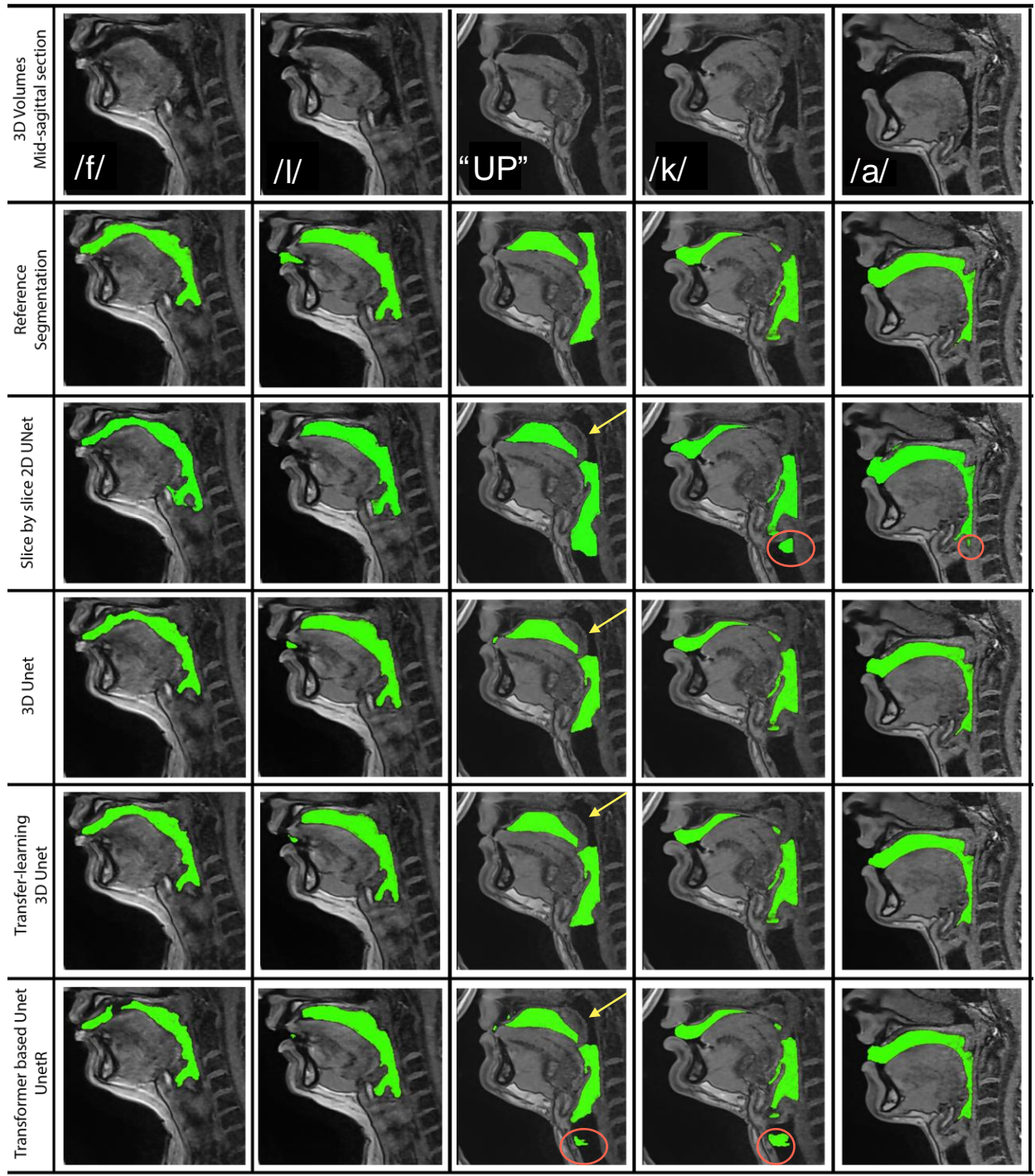

Figure 4: Mid-sagittal representations of the five test volumes used in the experiment are depicted in the first row. The second row presents the reference segmentation overlaid on a mid-sagittal slice for each volume. Subsequent rows display network segmentations overlaid on mid-sagittal slices, highlighting disjoint masks in columns two, three, and four. The specific sounds sustained during MRI scans (/f/, /l/, "UP", /k/, /a/) are labeled in the first row. Non-

*anatomical segmentations are highlighted in red ellipses, and missed segmentations are highlighted by yellow arrows.*

*Figure 5: 3D representations of the vocal tract airspace volumes generated from example speech postures in the test set. The specific sounds during MRI scans (/f/, /l/, "UP", /k/, /a/) are labelled in the first row. The second row shows the reference segmentation, while subsequent rows depict outputs from various neural network segmentations. The airway segmentation is highlighted in green, and non-anatomical segmentations marked with red circles. Differences in airway curvature among subjects and alterations during different vocalizations are observable. Differences between models are noted, particularly note the rough/leaky segmentations in the 2D slice-slice U-Net model near the epi-glottis region (yellow arrow), and missed segmentations in thin air-space region in the /k/ sound (magenta arrows).*

| Volume (Phoneme) | 3D Unet | | | UnetR | | | Unet3D-TL | | | 2D Slice by slice | | |
|---|---|---|---|---|---|---|---|---|---|---|---|---|
| | Dice Coefficient | HD Distance | SSIM Index | Dice Coefficient | HD Distance | SSIM Index | Dice Coefficient | HD Distance | SSIM Index | Dice Coefficient | HD Distance | SSIM Index |
| 1(/oe/) | 0.936 | 1 | 0.961 | 0.931 | 1.4 | 0.957 | 0.935 | 1.4 | 0.96 | 0.879 | 10.2 | 0.932 |
| 2(/a/) | 0.936 | 1.4 | 0.966 | 0.933 | 1.4 | 0.964 | 0.931 | 1 | 0.963 | 0.809 | 8 | 0.925 |
| 3(UP) | 0.871 | 18 | 0.957 | 0.867 | 18.5 | 0.949 | 0.884 | 16.6 | 0.959 | 0.81 | 18 | 0.938 |
| 4(/f/) | 0.899 | 2.2 | 0.951 | 0.877 | 3.3 | 0.943 | 0.906 | 2 | 0.952 | 0.821 | 13 | 0.92 |
| 5(/l/) | 0.912 | 2.4 | 0.952 | 0.901 | 4.5 | 0.949 | 0.901 | 2.2 | 0.949 | 0.848 | 7 | 0.93 |
| 6(/k/) | 0.929 | 1 | 0.964 | 0.908 | 23 | 0.953 | 0.924 | 1.4 | 0.963 | 0.885 | 2.8 | 0.948 |
| 7(/n/) | 0.909 | 4.2 | 0.963 | 0.914 | 1.4 | 0.965 | 0.89 | 2 | 0.956 | 0.774 | 18.2 | 0.917 |
| 8(/ʃ/) | 0.775 | 82 | 0.934 | 0.759 | 73.6 | 0.927 | 0.793 | 5 | 0.94 | 0.758 | 13.5 | 0.923 |
| Average | 0.896 ± 0.05 | 14 ± 28 | 0.956 ± 0.01 | 0.886 ± 0.06 | 15.9 ± 24.9 | 0.951 ± 0.01 | 0.896 ± 0.04 | 3.95 ± 5.2 | 0.955 ± 0.1 | 0.823 ± 0.05 | 11.33 ± 5.4 | 0.929 ± 0.01 |

*Table 2: Performance comparison of segmentation models (3D U-Net, U-NetR, U-Net 3D-Transfer Learning, and 2D slice-by-slice U-Net) across eight volumes. The metrics evaluated are Dice coefficient, HD distance, and SSIM Index. The table includes individual volume performance as well as the average performance and standard deviation for each metric. Higher Dice coefficient and SSIM Index values indicate better performance, while lower HD distance values indicate better performance. Notably, a trend in how posture affects segmentation quality is observed, particularly evident in Volume number 8 (/ʃ/) , where consistently lower Dice and higher HD distance are generated across all networks.*